# Utilizing Large Language Models for Named Entity Recognition in Traditional Chinese Medicine against COVID-19 Literature: Comparative Study

Xu Tong[1], PhD; Nina Smirnova[2], MSc; Sharmila Upadhyaya[2], MSc; Ran Yu[3], PhD; Jack H. Culbert[2], MSc; Chao Sun[4], PhD;  Wolfgang Otto[2], MSc; Philipp Mayr[2], PhD

[1] Institute of Basic Theory for Chinese Medicine, China Academy of Chinese Medical Sciences, Beijing, China
[2] GESIS—Leibniz Institute for the Social Sciences, Cologne, Germany
[3] University of Bonn, Bonn, Germany
[4] School of Traditional Chinese Medicine, Capital Medical University, Beijing, China

## Abstract

**Objective:** To explore and compare the performance of ChatGPT and other state-of-the-art LLMs on domain-specific NER tasks covering different entity types and domains in TCM against COVID-19 literature.

**Methods:** We established a dataset of 389 articles on TCM against COVID-19, and manually annotated 48 of them with 6 types of entities belonging to 3 domains as the ground truth, against which the NER performance of LLMs can be assessed. We then performed NER tasks for the 6 entity types using ChatGPT (GPT-3.5 and GPT-4) and 4 state-of-the-art BERT-based question-answering (QA) models (RoBERTa, MiniLM, PubMedBERT and SciBERT) without prior training on the specific task. A domain fine-tuned model (GSAP-NER) was also applied for a comprehensive comparison.

**Results:** The overall performance of LLMs varied significantly in exact match and fuzzy match. In the fuzzy match, ChatGPT surpassed BERT-based QA models in 5 out of 6 tasks, while in exact match, BERT-based QA models outperformed ChatGPT in 5 out of 6 tasks but with a smaller F-1 difference. GPT-4 showed a significant advantage over other models in fuzzy match, especially on the entity type of TCM formula and the Chinese patent drug (TFD) and ingredient (IG). Although GPT-4 outperformed BERT-based models on entity type of herb, target, and research method (RM, 0.443), none of the F-1 scores exceeded 0.5. GSAP-NER, outperformed GPT-4 in terms of F-1 by a slight margin on RM. ChatGPT achieved considerably higher recalls than precisions, particularly in the fuzzy match.

**Conclusions:** The NER performance of LLMs is highly dependent on the entity type, and their performance varies across application scenarios. ChatGPT could be a good choice for scenarios where high recall is favored. However, for knowledge acquisition in rigorous scenarios, neither ChatGPT nor BERT-based QA models are off-the-shelf tools for professional practitioners.

# Introduction

## Background

Traditional Chinese Medicine (TCM) has emerged as a valuable approach in the fight against COVID-19 [1]. Multiple studies have revealed the safety and effectiveness of TCM in the treatment of COVID-19 [2–4]. By combining herbal formulas, acupuncture, moxibustion, and other therapeutic techniques, TCM offers a multifaceted approach to support and plays a vital role in the prevention and treatment of the COVID-19 pandemic during the past three years [4, 5]. Understanding the foundational principles and therapeutic mechanism of TCM from current literature is essential to better apply TCM in response to such a global health crisis, and perhaps other similar health crises in the future. TCM against COVID-19 publications have resulted in a relatively niche and highly specialized dataset containing a vast quantity of cross-domain entities encapsulating rich information relevant to TCM, network pharmacology, general science domain, etc., which needs further exploration. Therefore, it is crucial to find a suitable method for extracting this valuable information automatically. With the aid of named entity recognition (NER), valuable information can be rapidly extracted from unstructured texts, laying the foundation for the construction of TCM knowledge graphs, so that researchers can apply advanced AI methods to gain new insights into TCM for similar health crises in the future.

Named entity recognition (NER) approaches have historically been trained in a fully supervised manner and rely mostly on manually annotated corpora of sufficient size, which are not easily obtainable due to high costs and time demands. Despite significant progress in developing supervised NER models for multiple languages and domains, scaling to emerging and low-resource domains remains challenging, due to the costly actuality of acquiring training data [6]. For the aforementioned TCM against covid-19 literature, there is currently no manually well-annotated training data that can be directly relied on because of the novel, niche and specialized nature of such data, which increases the challenge and difficulty of applying supervised learning NER. Meanwhile, the rise of large language models (LLMs) has stimulated a new research boom as they have shown powerful capabilities to solve a wide variety of natural Language Processing (NLP) tasks, especially for the zero-shot or few-shot scenarios [7]. As one of the most widely used LLMs, ChatGPT has attracted massive public and academic attention. Besides its conversational skills, ChatGPT has exhibited remarkable performance in many NLP tasks on various domain-specific datasets, such as annotation and information extraction in news media data [8-10], NER and reasoning in health care data [11, 12], and text analytics in historical and financial texts [13, 14], without any fine-tuning or with minimal fine-tuning using a very small amount of data [12]. For research in the NLP of textual content, a focus of current research is on exploring the capabilities of ChatGPT on entity extraction tasks. However, there are few results on how ChatGPT performs on NER tasks on low-resource and niche domain datasets, particularly in the field of TCM against COVID-19 literature.

Therefore, we aim to explore the performance of ChatGPT on NER tasks in TCM against COVID-19 literature over different entity types and different domains, and simultaneously compare ChatGPT with other state-of-the-art large language models. The approaches we used for comparison are BERT (Bidirectional Encoder Representations from Transformers) [15] based question answering (QA) approaches. The reasons for choosing these models are as follows: (1) BERT is another groundbreaking language model that has revolutionized NLP tasks. Its contextual understanding allows it to decipher the nuances of language, making it highly adaptable to various domains. (2) Previous studies showed that QA models can leverage knowledge learned from question-answering data to improve low-resource NER performance [16, 17], making it suitable for our TCM against the COVID-19 dataset, which has the characteristics of low-resource and lack of annotated data.

# Related work

## LLMs and ChatGPT

As a well-known achievement of advanced artificial intelligence, LLMs have demonstrated impressive capabilities in tackling diverse NLP tasks. Nowadays, numerous large-scale models, notably ChatGPT [18, 19], BERT [15], T5 [20], and PaLM [21], have been proposed, featuring millions to over 100 billion parameters. LLMs exhibited remarkable emergent abilities that facilitate adept performance in zero-shot and few-shot situations. ChatGPT is one of the most prominent LLMs. The release of GPT-3.5 and GPT-4 on 30th November 2022 and 14th March 2023 respectively has led to an increase in research investigating ChatGPT's potential for NER across various tasks and domains. Nonetheless, the reported performances of ChatGPT differed significantly in different studies [7, 22]. Some studies confirmed ChatGPT's impressive zero- and few-shot capabilities in NER [23], achieving performance similar to a pre-trained language model [24] and even surpassing the full-shot models on several datasets [25]. However, in other studies, ChatGPT encountered difficulties and lagged behind previous fine-tuning methods in NER in specific domains, like historic documents [13], gene association database [26], financial text [14], and even general NER dataset [27]. In the clinical and biomedical field, ChatGPT also exhibits a capacity to identify and classify medical entities [28, 29], showcasing its potential as a valuable tool for NER in healthcare-related texts. Although in another study where ChatGPT achieved poor performance for the biomedical NER task [30], there are still studies claiming the great potential and interest of ChatGPT for clinical and biomedical NER tasks as it does not require any annotation [12,31], mitigating ChatGPT's limitation and enabling it to excel across an even wider spectrum of medical domain-specific NER tasks. Recently, multiple research endeavors into NER in TCM data have been carried out, including ancient Chinese medicine books and clinical medical records, which possess characteristics such as a large number of entities like formulae, herbs [32], multi-ingredient, and multi-targets [33]. But all of these research are focused on supervised learning [32, 34], unsupervised learning and available annotated datasets are still scarce. To date, to the best of our knowledge, no studies have utilized ChatGPT for the aforementioned multiple entity recognition in TCM literature.

## BERT-based models

Bidirectional Encoder Representations Transformers (BERT) [15] is a transformer-based language representation model that aims to capture contextualized embedding of words. These embedding serve as a numerical representation of textual data in a vector space, mapping words with similar meanings to similar vectors and therefore generating comparable representations. Succeeding the proposal of BERT, several modified BERT variants have been proposed. These variations are either minor or significant modifications to the architecture, pre-training techniques or training data to improve BERT's performance for various tasks.

In the present paper, several variants of the BERT model were used to perform NER tasks on TCM against the COVID-19 literature. RoBERTa [35] is an optimization of the BERT model, which was evaluated on different benchmarks and demonstrated massive improvements over reported BERT performance [36]. A RoBERTa-based question-answering (QA) approach [37] has been used to identify adverse drug events, in which the QA model was applied as the last step of the extraction pipeline, preceded by NER and the classification module, showing state-of-the-art performance. MiniLM [38] is a novel approach to distill larger models, such as BERT, into smaller models that maintain high accuracy while being much faster than larger models. SciBERT [39] was proposed to address the shortcomings of BERT when working with scientific datasets. SciBERT has been trained on a large multi-domain corpus of scientific publications and surpasses the performance of BERT for various tasks such as sentence classification, dependency parsing, and sequence tagging in the scientific domain. PubMedBERT [40] is another variation of BERT that was trained on scientific articles from the biomedical domain. It has achieved state-of-the-art results for NLP tasks in the biomedical field [41].

Question-answering-based NER

With the rise of the BERT-based models, state-of-the-art results have been achieved in various NLP tasks, among which NER has been studied extensively. In particular for the NER task on niche and specific domain data where annotation requires human experts and costs more time, various works have been carried out to resolve the NER problem using zero-shot and few-shot learning. One of the prominent approaches is using a QA pipeline, where questions represent the desired entity in a document and extracted answers include sections of input text that contain such entities. The QaNER model, which uses extractive QA models to identify desired entities, was introduced in 2022 [16]. Extractive QA models extract answers from the text as a substring, which is perfectly suited to perform an NER task. SQuAD [42] and SQuAD 2.0[43] datasets were used to train a QA model on top of BERT. The QA models and different prompts were tested on different benchmark datasets and demonstrated good performance for the NER task. Another study applied a QA-based approach to identify funding information in scientific acknowledgements texts, which showed good performance in this specific NER task [17]. Several transformer-based models trained on the SQuAD 2.0 datasets [43] and various prompts were tested on a sample of scientific texts retrieved from Crossref [44], among which RoBERTa and MiniLM models showed the highest performance.

# Methods

## Data collection

In this paper, we focus on investigating the performance of different LLMs on NER in TCM against COVID-19 literature. Hence we select our experimental data from existing high-quality research data corpus for medical-related research to avoid introducing noise into the domain focused analysis. Specifically, we used two sources of literature written in English of TCM against COVID-19: COVID-19 Open Research Dataset (CORD-19) [45] and PubMed-sourced articles.

(1) CORD-19. CORD-19 [45] is a dataset containing more than 1,000,000 scholarly articles, focusing on COVID-19, SARS-CoV-2, and related coronaviruses. It's publicly available, enabling researchers to employ advanced NLP and AI methods for fresh insights in the fight against COVID-19. The dataset we used was the version updated in June 2022. We used 49 keywords of "Traditional Chinese medicine", "COVID-19" and their corresponding synonyms as search terms. These search terms were derived from PubMed's MeSH Database, and are entry terms under the MeSH term for each keyword to ensure a complete and accurate search. After deduplication and manual removal of irrelevant literature, we obtained 344 articles.

(2) PubMed-sourced articles. We searched PubMed to get more articles regarding TCM against COVID-19 published between the 1st of July 2022 and the 23rd of May 2023, to fill the gap after June 2022 of CORD-19. The search strategy we used was "Medicine, Chinese Traditional [MeSH]" and "COVID-19 [MeSH]", ensuring that the search terms are the same as those we applied to CORD-19. After deduplication and manual removal of irrelevant literature, we obtained 61 articles.

After filtering out 16 duplicates between the two data collections, we obtained a total of 389 articles as our dataset. (We made this dataset openly available, more information can be found in Data availability section.) We extracted titles and abstracts from these articles to use as research data in our analysis. The average length of the titles and abstracts was 20 and 259 words, respectively.

## Tasks

We aim to explore the performance of different state-of-the-art LLMs on the task of NER for 6 entity types. We first defined 6 tasks each covering one entity type, belonging to 3 different domains, as listed below. Table 1 provides the entity type description and examples of each type.

Domain 1: TCM domain
Task 1: The TCM formula and the Chinese patent drug (TFD)
Task 2: Herb (HB)

Domain 2: Network pharmacology domain
Task 3: Ingredient (IG)
Task 4: Target (TG)

Domain 3: General science domain
Task 5: Study design (SD)
Task 6: Research method (RM)

 **Table 1.** The entity type description and examples.

| Domain | Task (Entity type) | Entity type description | Entity examples |
|---|---|---|---|
| 1 | **TFD** | A combination of herbs that are prescribed together to treat a specific health condition or set of symptoms. | Xuebijing, Lianhua Qingwen |
|  | **HB** | A natural plant or plant part that is used for medicinal purposes. | Asari Radix et Rhizoma, Semen Sojae Preparatum |
| 2 | **IG** | A substance contained in the herbs, TCM formulas and Chinese patent drugs registered in the Pharmacopoeia of the People's Republic of China or other literates. | procyanidin B1 eriodictyol, quercetin |
|  | **TG** | A biological molecule or a cellular process that plays a key role in a disease or biological pathway. | SARS-COV-2 3CL, ACE2 |
| 3 | **SD** | The systematic and overall plan or strategy that is used to conduct the study. | systematic review, cross-sectional study |
|  | **RM** | The specific approach or technique used in the study process of collecting data, analyzing data, etc. | KEGG enrichment analysis, factor analysis |

## Manual annotation and ground truth

We created a codebook with specific definitions and guidelines for each task, and trained a team of two TCM experts to annotate the titles and abstracts of 48 articles randomly selected from our dataset. For all 6 tasks mentioned above, the annotators were given the same set of instructions in the codebook. Then, the annotators were asked to annotate the titles and abstracts task by task independently. Subsequently, the two experts discussed and solved the disagreements. We relied on the annotation to construct a ground truth against which the performance of ChatGPT and other models can be assessed. Table 2 presents the statistics of the ground truth.

**Table 2.** Statistics of the ground truth.

| Domain | Task (Entity type) | Number of distinct entities | Number of articles containing entities |
|---|---|---|---|
| 1 | TFD | 21 | 23 |
|   | HB | 11 | 3 |
| 2 | IG | 31 | 9 |
|   | TG | 60 | 12 |
| 3 | SD | 23 | 22 |
|   | RM | 50 | 28 |

# Models

In this section, we describe the models used in our research, namely ChatGPT (GPT-3.5 and GPT-4) and 4 different BERT-based QA models (RoBERTa, MiniLM, PubMedBERT and SciBERT). All the above-mentioned models produce NER results on the 48 titles and abstracts without any prior training on the specific tasks. For a comprehensive comparison, we also used GSAP-NER, which is a domain fine-tuned model specifically developed for research methods entity extraction, to compare the performance on task 6 (RM) with above LLMs.

ChatGPT

ChatGPT, a well-known LLM from OpenAI, is capable of identifying and extracting named entities from user input, while considering the context of the conversation through the use of appropriate prompts. Recent studies have shown that careful prompt design, also known as prompt engineering, is crucial to achieve good performance [24]. In our case, we developed 3 different prompts for each task that would effectively activate ChatGPT's NER capabilities. Table 3 presents the prompts we used for "herbs" as an example. Prompt #1 simply describes the tasks; prompt #2 gives a role of the NER system to ChatGPT allowing it to better understand the tasks and deliver the desired results; prompt #3, in addition to prompt #2, provides definitions and examples of herbs. We prepended each prompt to a title and abstract to form a query to ChatGPT.

**Table 3.** Different prompts for ChatGPT using herbs as an example.

| Prompt | Content |
| --- | --- |
| **#1** | Please extract the herbs from the following text and output the herbs in a list without any other introduction. |
| **#2** | You are a smart and intelligent Named Entity Recognition (NER) system, which is very familiar with named entities in the field of Chinese medicine. Please extract the herbs from this following text and output the herbs in a list without any other introduction. |
| **#3** | You are a smart and intelligent Named Entity Recognition (NER) system, which is very familiar with named entities in the field of traditional Chinese medicine. Herb refers to a natural plant or plant part that is used for medicinal purposes, such as Asari Radix et Rhizoma, Semen Sojae Preparatum. Please extract the herbs from this following text and output the herbs in a list without any other introduction. |

BERT based question-answering model

We used 4 different pre-trained BERT-based models fine-tuned for the QA task:
- RoBERTa [35]
- MiniLM [38]
- SciBERT [39]
- PubMedBERT [40]

For this work, we selected the above pre-trained BERT-based models, which were further fine-tuned for QA tasks on the SQuAD [42] and COVID-QA [46] datasets. To perform these QA models for our tasks, we developed different questions corresponding to each task. Table 3 shows the designed questions for recognizing entities of type herbs as an example.

The QA pipelines require the input consists of a context C and a question Q, and then outputs text spans from context C as an answer A to the question Q. In our case, the context C is the concatenation of 48 abstracts and titles, question Q is the set of questions we developed for each entity type as illustrated in Table 4. The answer A refers to the text spans extracted from context C in response to the question Q, i.e., the desired entities.

**Table 4.** Questions used for the BERT-based question-answering models to extract entities of type herbs.

| Number | Questions |
| --- | --- |
| #1 | What is the herb in the text? |
| #2 | What are the herbs in the text? |
| #3 | What are the medicinal herbs in the text? |
| #4 | What are the Chinese medicinal herbs in the text? |
| #5 | What are the medicinal plants in the text? |
| #6 | What herbs (medicinal herbs/ Chinese medicinal herbs/ medicinal plants) are |
| #7 | What herbs (medicinal herbs/ Chinese medicinal herbs/ medicinal plants) |
| #8 | What's the name of the herb in the text? |
| #9 | What are the names of the herbs (medicinal herbs/ Chinese medicinal herbs/ |
| #10 | Which is the herb in the text? |
| #11 | Which herbs (medicinal herbs/ Chinese medicinal herbs/ medicinal plants) |

GSAP-NER

For a more comprehensive comparison, we further applied GSAP-NER [47,48] on task 6 (RM) to compare the performance between the domain fine-tuned language model and LLMs. GSAP-NER is a collection of the state-of-the-art methods trained on a corpus of domain-specific scholarly articles to extract scholarly entities and concepts. We used the best-performing baseline model following [48], which is a fine-tuned SciDeBERTa-CS language model [49]. Similar to BERT-based QA models, the output is text spans identified in the titles and abstracts for each entity type.

# Evaluation

## Evaluation Metrics

To compare the performance of different models, we choose the commonly used evaluation metrics [50, 51] for NER tasks, namely, precision (P), recall (R), and F-1 score.

The formulas for P, R and F-1 scores are as follows:

$$P = \frac{TP}{(TP + FP)}$$
$$R = \frac{TP}{(TP + FN)}$$
$$F-1 = 2 \times \frac{P \times R}{(P + R)}$$

True positive (TP) is an outcome where the model correctly predicts the positive class, i.e., predicted entity that corresponds to the ground truth entity. False positive (FP) is incorrectly predicted positive class, i.e., predicted entity that does not correspond to any ground truth entity. False negative (FN) is an incorrectly predicted negative class, i.e., a ground truth entity, which does not correspond with any predicted entities. Therefore, precision is a ratio of true positives to true and false positives. Recall is a ratio of true positives to true positives and true negatives. The F-1 score is a harmonic mean of precision and recall.

Matching method

We address the task of NER in TCM against COVID-19 literature, however, in different application scenarios, the specific goal of the NER task may vary. For instance, a domain expert may be interested in formulas containing a specific herb, while for a domain novice, it may be important to get annotations with higher recall to have a better overview of the topic. Hence, we further consider two varied task setups, i.e., exact match and fuzzy match, to have better coverage of different application scenarios.

Exact match considers only strings containing exactly the same tokens in exactly the same order as a match, while fuzzy match considers semantically similar but not necessarily identical literals as matches. Thus, two strings in Example 1 will be identified as not match in the exact match setting but can be identified as a match in the fuzzy match setting.

*Example 1:* 'pudilanxiaoyan oral liquid pdl', 'pudilanxiaoyan oral liquid'

For the fuzzy match evaluation, we use the python library fuzzywuzzy [52] with a threshold of 60%, i.e., strings with a similarity score higher than 60% were marked as a match. For example, the similarity score between two strings in Example 1 is 93%.

Post-processing

For this study, the ground truth entities and predicted entities were processed prior to the evaluation. All entities were tokenized, lowercased and stemmed. All punctuation marks, non-ASCII characters, and stop words [53] were removed. Tokenization in NLP is a process of splitting a sentence or paragraph into smaller units, in our case, words. Stemming reduces a word to its root. Therefore, the string in Example 2 after cleaning was transformed into the string in Example 3.

Example 2: 'Pudilanxiaoyan oral liquid (PDL)'

Example 3: 'pudilanxiaoyan oral liquid pdl'

The aforementioned cleaning and stemming pipeline was adopted for both ChatGPT and BERT based QA models. As the BERT-based QA models we used are extractive and hence the answers were extracted from the input text itself i.e., no noise. However, in the case of the ChatGPT, the results are generative, i.e., responses can be elaborated as full sentences. To make the results from different models comparable, we further performed post-processing steps to extract NER results from the responses generated from ChatGPT, which are:

1. Removing responses including negations
   Example: 'Sorry, there are no TCM mentioned in this text'
2. Removing descriptive text to only keep the recognized entities from the sentence
   Example: 'The herb used in this text is'
3. Removing the entity type names
   Example: 'tcm formula', 'Chinese patent drug', 'target', 'ingredient'

# Results

## Overall results

As shown in Table 2-3, different prompts were crafted for ChatGPT and different questions were developed for BERT-based QA models for each task. For better readability, in Table 5, we present only the results of the best-performing question/prompt as the indicator of the potential of each model on the NER tasks we focus on, to facilitate a fair comparison between models.

ChatGPT and BERT-based QA models exhibited varied performance across the different types of entities in both exact match and fuzzy match, shown in table 5. Figure 1-2 show a clear performance difference of each model on each task in terms of F-1 score in the fuzzy match and exact match, respectively. BERT-based QA models outperformed ChatGPT in 5 tasks in the exact match, i.e., PubMedBERT performed best in the task of HB and SD, RoBERTa performed best in the task of IG and TG, MiniLM performed best in the task of RM, while GPT-4 outperformed others in the task of TFD. But in terms of F-1 in fuzzy match, GPT-4 outperformed BERT-based QA models in 5 tasks of TFD, HB, IG, TG and RM, and RoBERTa outperformed ChatGPT in the task of SD. The domain-fine-tuned tool GSAP-NER outperformed both ChatGPT and BERT-based QA models in the task of RM. Across all tasks, GPT-4 achieved higher F-1 scores than GPT-3.5.

**Table 5**. Summary of model performance by each task. The highest F-1 score on each task is in bold.

| Models | Task | Exact match | | | Fuzzy match | | |
|---|---|---|---|---|---|---|---|
| | | Precision | Recall | F-1 | Precision | Recall | F-1 |
| **RoBERTa** | TFD | 0.152 | 0.250 | 0.189 | 0.283 | 0.500 | 0.361 |
| | HB | 0.023 | 0.091 | 0.036 | 0.064 | 0.273 | 0.103 |
| | IG | 0.133 | 0.176 | **0.152** | 0.156 | 0.206 | 0.177 |
| | TG | 0.067 | 0.042 | **0.052** | 0.156 | 0.097 | 0.120 |
| | SD | 0.109 | 0.161 | 0.130 | 0.313 | 0.500 | **0.385** |
| | RM | 0.229 | 0.141 | 0.175 | 0.468 | 0.286 | 0.355 |
| **MiniLM** | TFD | 0.083 | 0.138 | 0.104 | 0.208 | 0.370 | 0.267 |
| | HB | 0.021 | 0.091 | 0.034 | 0.042 | 0.182 | 0.068 |
| | IG | 0.042 | 0.065 | 0.051 | 0.146 | 0.200 | 0.169 |
| | TG | 0.021 | 0.014 | 0.017 | 0.042 | 0.027 | 0.033 |
| | SD | 0.061 | 0.097 | 0.075 | 0.163 | 0.267 | 0.203 |
| | RM | 0.271 | 0.160 | **0.202** | 0.458 | 0.278 | 0.346 |
| **PubMedBERT** | TFD | 0.176 | 0.200 | 0.188 | 0.353 | 0.414 | 0.381 |
| | HB | 0.050 | 0.100 | **0.067** | 0.150 | 0.273 | 0.194 |
| | IG | 0.152 | 0.143 | 0.147 | 0.455 | 0.417 | 0.435 |
| | TG | 0.065 | 0.023 | 0.034 | 0.250 | 0.103 | 0.146 |
| | SD | 0.139 | 0.147 | **0.143** | 0.306 | 0.324 | 0.314 |
| | RM | 0.179 | 0.084 | 0.115 | 0.359 | 0.169 | 0.230 |
| **SciBERT** | TFD | 0.085 | 0.161 | 0.111 | 0.167 | 0.393 | 0.234 |
| | HB | 0.036 | 0.182 | 0.061 | 0.068 | 0.364 | 0.114 |
| | IG | 0.066 | 0.143 | 0.090 | 0.171 | 0.361 | 0.232 |
| | TG | 0.034 | 0.023 | 0.027 | 0.119 | 0.094 | 0.105 |
| | SD | 0.100 | 0.171 | 0.126 | 0.233 | 0.400 | 0.295 |
| | RM | 0.035 | 0.025 | 0.029 | 0.293 | 0.207 | 0.243 |
| **GPT-3.5** | TFD | 0.172 | 0.345 | 0.230 | 0.483 | 0.966 | 0.644 |
| | HB | 0.019 | 0.182 | 0.035 | 0.106 | 1.000 | 0.191 |
| | IG | 0.022 | 0.143 | 0.038 | 0.151 | 0.971 | 0.262 |
| | TG | 0.018 | 0.092 | 0.031 | 0.171 | 0.890 | 0.286 |
| | SD | 0.042 | 0.188 | 0.069 | 0.182 | 0.867 | 0.301 |
| | RM | 0.051 | 0.200 | 0.082 | 0.195 | 0.867 | 0.319 |
| **GPT-4** | TFD | 0.276 | 0.276 | **0.276** | 0.800 | 0.828 | **0.814** |
| | HB | 0.035 | 0.182 | 0.059 | 0.193 | 1.000 | **0.324** |
| | IG | 0.097 | 0.167 | 0.122 | 0.531 | 0.971 | **0.687** |
| | TG | 0.023 | 0.103 | 0.037 | 0.193 | 0.919 | **0.319** |
| | SD | 0.046 | 0.212 | 0.075 | 0.150 | 0.719 | 0.249 |
| | RM | 0.071 | 0.207 | 0.106 | 0.292 | 0.918 | 0.443 |
| **GSAP-NER** | RM | 0.134 | 0.105 | 0.118 | 0.537 | 0.387 | **0.450** |

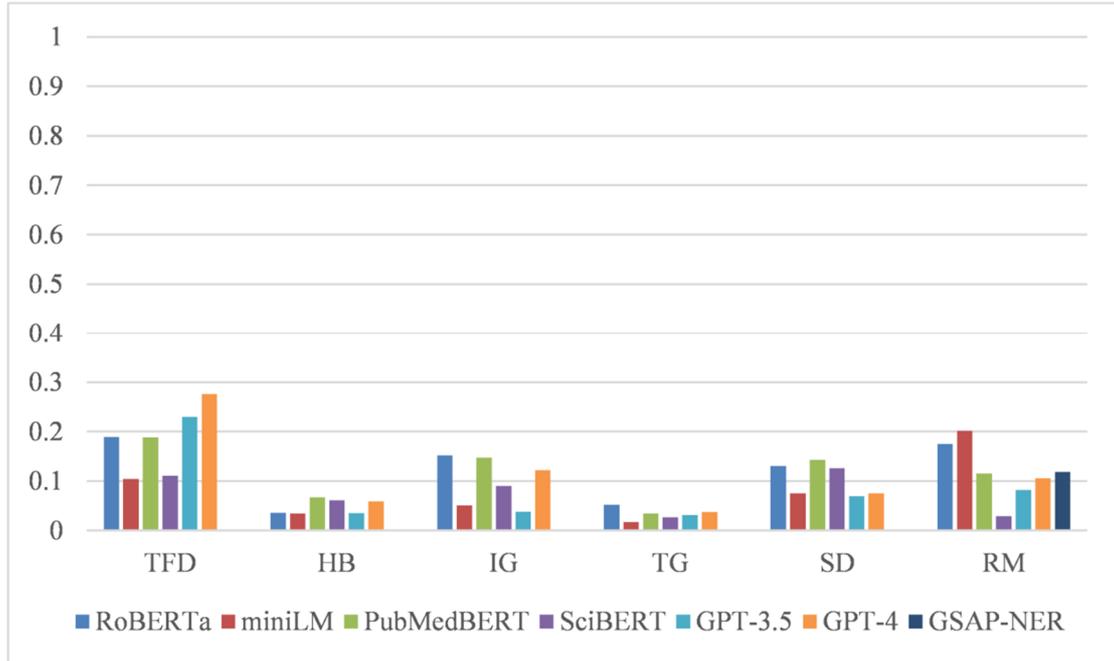

**Figure 1.** The performance of each model on each task in terms of F-1 score in the exact match.

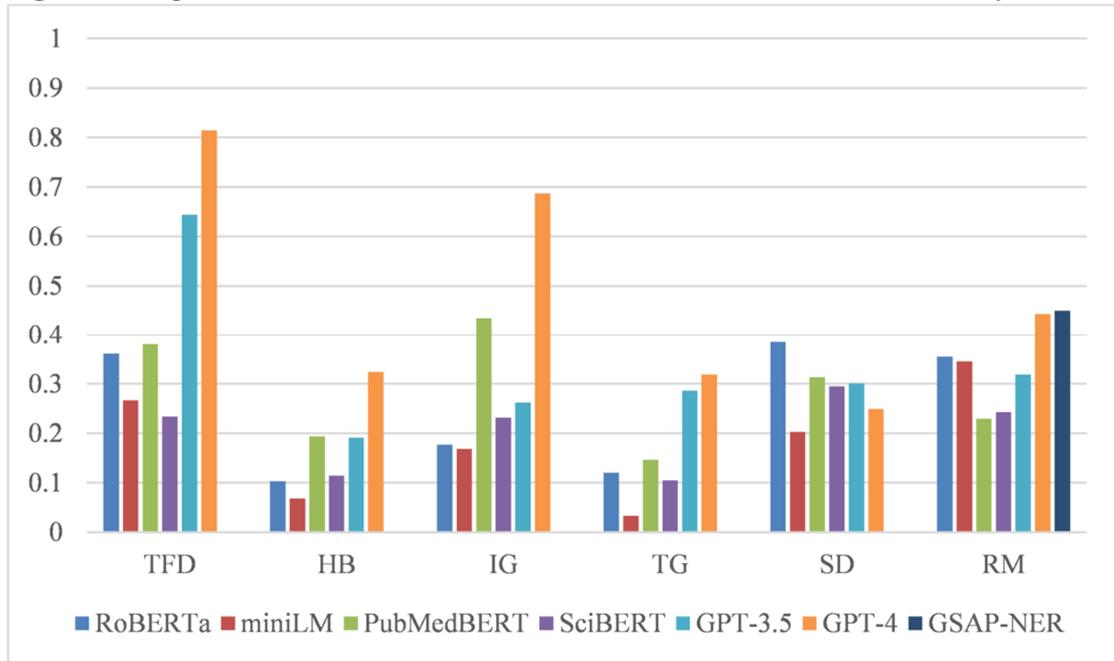

**Figure 2**. The performance of each model on each task in terms of F-1 score in the fuzzy match.

In order to know if there was a substantial performance gap between ChatGPT and BERT-based QA models on each task, we further calculated the difference in F-1 scores between ChatGPT and BERT-based QA models (Table 6-7). For instance, in the fuzzy match where ChatGPT outperformed BERT-based QA models in 5 tasks, the F-1 score of ChatGPT (GPT-4) on the TFD task was 0.814, and the highest F-1 score from BERT-based QA models on TFD was 0.381 (PubMedBERT), the performance gap between them was 0.433 (53.2%).

**Table 6.** The performance gap between BERT-based QA models and ChatGPT in terms of F-1 score in the exact match.

| Task | F-1 score in the exact match | | | | △ F-1 (%) |
|------|------|------|------|------|------|
|      | BERT-based QA | | ChatGPT | | |
| HB | PubMedBERT | 0.067 | GPT-4 | 0.059 | 0.008 (11.9%) |
| IG | RoBERTa | 0.152 | GPT-4 | 0.122 | 0.03 (19.7%) |
| TG | RoBERTa | 0.052 | GPT-4 | 0.037 | 0.015 (28.8%) |
| SD | PubMedBERT | 0.143 | GPT-4 | 0.075 | 0.068 (47.6%) |
| RM | MiniLM | 0.202 | GPT-4 | 0.106 | 0.096 (47.5%) |

**Table 7.** The performance gap between ChatGPT and BERT-based QA models in terms of F-1 score in the fuzzy match.

| Task | F-1 score in the fuzzy match | | | | △ F-1 (%) |
|------|------|------|------|------|------|
|      | ChatGPT | | BERT-based QA | | |
| **TFD** | GPT-4 | 0.814 | PubMedBERT | 0.381 | 0.433 (53.2%) |
| **HB** | GPT-4 | 0.324 | PubMedBERT | 0.194 | 0.13 (40.1%) |
| **IG** | GPT-4 | 0.687 | PubMedBERT | 0.435 | 0.252 (36.7%) |
| **TG** | GPT-4 | 0.391 | PubMedBERT | 0.146 | 0.245 (62.7%) |
| **RM** | GPT-4 | 0.443 | RoBERTa | 0.355 | 0.088 (19.9%) |

# Discussion

## Principal findings

We investigated and compared the NER capabilities of ChatGPT (GPT-3.5 and GPT-4) along with BERT-based QA models (RoBERTa, MiniLM, PubMedBERT and SciBERT) on a low-resource domain dataset, i.e., TCM against COVID-19 literature dataset. It is a relatively niche and highly specialized dataset containing multiple entities encapsulating rich information relevant to TCM, network pharmacology, and general science. Consequently, it also lacks publicly available and reliable annotations，which we are contributing to through this work. The overall performance of the applied two types of language models varied significantly in both exact match and fuzzy match. In terms of F-1 score, BERT-based QA models surpassed ChatGPT in 5 different tasks (HB, IG, TG, SD, and RM) in the exact match, whereas ChatGPT outperformed BERT-based QA models in 5 different tasks (TFD, HB, IG, TG, and RM) in the fuzzy match. One possible reason for ChatGPT performing better in the fuzzy match while BERT-based QA models performing better in the exact match may be that, ChatGPT is a generative language model, which can generate similar semantic entities by understanding the input text and prompts, but the generated entities would not be exactly the same as the entities in the input text. While the way of BERT-based QA models to resolve the NER task is by extracting entities from input text as answers. Furthermore, the performance difference between ChatGPT and BERT-based QA models on each task was notable. Specifically, in the fuzzy match ChatGPT surpassed the BERT-based QA models, with GPT-4 achieving a higher F-1 score gap by 0.252 (36.7%) and 0.433 (53.2%) on IG and TDF, respectively. When the BERT-based QA models outperformed ChatGPT in exact match, the F-1 score gap achieved by the BERT-based QA models was only marginally higher than ChatGPT by no more than 0.1 (Table 6). However, it is important to note that all F-1 scores in the exact match were quite low, none of which exceeded 0.5 (Table 6, 7). That is to say, neither ChatGPT nor the BERT-based QA models are able to achieve satisfactory NER performance on our dataset in the exact match setting.

To more accurately reflect the performance of LLMs, especially ChatGPT, which tend to identify longer spans of text to get closer to human responses, we focus on the results of the fuzzy match. In the fuzzy match, GPT-4 outperformed BERT-based QA models in 5 tasks of TFD, HB, IG, TG, and RM. On the task of TFD and IG, GPT-4 performed well and achieved F-1 scores of 0.814 and 0.689 respectively, which were comparable to the results obtained by ChatGPT in other relevant medical (highest F-1 of 0.776) [12] or clinical NER research (highest F-1 score of 0.677) [29]. But on the other 3 tasks of HB, TG and RM, GPT-4 achieved F-1 scores of around 0.4 and neither was higher than 0.5 (0.324, 0.319 and 0.443, respectively). We defined these 6 types of entities belonging to 3 different domains (Table 1), and we observed performance variations in different domains. In domain 3 (the general science domain), RoBERTa and GPT-4 attained comparable F-1 scores on SD and RM tasks (0.385 vs 0.443). While in domain 1 (TCM domain) ChatGPT's performance varied immensely on the two tasks of HB and TFD (0.814 vs 0.324). A reason may be that TCM practitioners tend to use TFD rather than individual herbs in clinical practice, many well-known TFDs were reported on the internet, especially during the time of COVID-19 when patent drugs came into the market and were available for more audience. Hence, if ChatGPT has been trained on the TCM domain corpus, it is highly likely to have been exposed to the TFD-related data. Furthermore, the TFDs usually have unique names written in Chinese pinyin and are typically followed by nouns like decoction, formula, powder, etc. So the characteristics of TFD names are easy for ChatGPT to understand and recognize. In contrast, it is more difficult to identify the herbs as our text are all in English in which the names of herbs are either in English or in Latin, and in some texts herb names are the plant names, while in others are the names for medicinal parts of the plants. In domain 2 (network pharmacology domain), the performance of GPT-4 varied significantly between IG and TG (0.687 vs 0.319) as well. NER performance of GPT-4 on TG is particularly challenging, exhibiting the lowest performance among all models. As a relatively novel concept in network pharmacology [54], TG mostly appears in specialized medical or bioinformatics research and is not commonly used in daily contexts. So we discovered that the disparity in performance of each model is not solely related to the domain, but more significantly, connected to the type of the particular entity.

As aforementioned, the GSAP-NER, a domain-tuned model specifically developed for research methods entity extraction, was also used to compare the performance on task 6 (RM). Although it had not been retrained on our dataset, GSAP-NER outperformed GPT-4 in terms of F-1 in the fuzzy match by a slight margin (0.45 vs 0.433), which is similar to the results obtained in another study [14], demonstrating that ChatGPT did not perform as well as domain-specific trained models. Unlike other studies that have argued that ChatGPT performs poorly on NER tasks (with F-1 scores of 0.072 and 0.379) [30] (0.174 to 0.44) [27], we would conclude that ChatGPT's NER performance is highly dependent on entity type.

Meanwhile, we observed from Table 5 that the recalls were much higher than precisions for ChatGPT, especially in the fuzzy match. That is to say, ChatGPT was able to achieve a higher coverage rather than a higher accuracy of positive matches. Thus, ChatGPT has an advantage in certain application scenarios with the requirement for high recall. For instance, in practical scenarios associated with TCM, from a user standpoint, ChatGPT might be suitable for domain novices who are willing to broaden their understanding in the area, particularly concerning entities like TCM formula, patented drugs, and active ingredients, providing them assistance in obtaining a more extensive overview of the subject matter. However, it should be noted that for all models, the recall (no higher than 0.3) and the F-1 values were all relatively low (no higher than 0.5) in the exact match, so that neither ChatGPT nor the BERT-based QA models are currently off-the-shelf tools for professional practitioners in rigorous knowledge acquisition scenarios. Thus, we would conclude that LLM's feasibility may depend on the application scenarios. Besides, from the perspective of methodology, the BERT-based QA models (RoBERTa, PubMedBERT, MiniLM) outperformed ChatGPT on several tasks in exact match. Despite the unsatisfactory F-1 scores, the BERT-based models are open-source and reportedly possess the ability to maintain high accuracy in various NLP tasks while being faster than larger models, making them more accessible for scientific research and hence worth further exploration.

# Limitations

In this study we focus on the comparison of the performance of LLMs on NER tasks in TCM against COVID-19 literature, so we discuss the best-performing configuration of each model in terms of different prompts and questions as the indicator of their potential. A more detailed analysis of the impact of different prompts and questions is not within the scope of this work. However, it is intuitive that the model performances can be further improved with more iterations in the prompt and question design, which poses a direction for future work.

Due to the niche and specialized nature of our dataset, and the lack of available annotation, the size of our study data was relatively compact. We have made our entire dataset and codebook available in the hope of assisting NER and NLP researchers working in low-resourced domains such as TCM. It is noteworthy that named entity annotation adheres to clearly defined rules to detect the nature and boundaries of universal and domain-specific entity types. While universal entity types, such as person, location, and organization, have stable definitions, the entity definitions and types in our domain-specific dataset, namely TCM against COVID-19, are not universal and may differ from other guidelines and research.

# Conclusions

Our findings reveal that the NER performance of LLMs is highly dependent on the entity type, and their performance varies across application scenarios. ChatGPT could be suitable for scenarios where high recall is expected, e.g. domain novices who are willing to broaden their understanding in the area, particularly regarding entity types like TCM formula, patented drugs, and active ingredients, providing them assistance in obtaining a more extensive overview of the subject matter. However, for rigorous knowledge acquisition scenarios, neither ChatGPT nor BERT-based QA models represent an off-the-shelf tool for professional practitioners. Additionally, some BERT-based QA models are open-source, which makes them more appropriate for scientific research.

# Acknowledgements

This work was supported by the Scientific and Technological Innovation Project of China Academy of Chinese Medical Sciences (CI2021A00409), the National Natural Science Foundation of China (82305439), the Capital Health Development Research Programme for Young Excellence (SF 2022-4-4282), Special fund of young scientific and technological talents of China Academy of Chinese Medical Sciences (ZZ16-YQ-053), the China Scholarship Council (202005350015), Deutsche Forschungsgemeinschaft (DFG) (MA3964/7-2, Pollux) and (460037581, BERD@NFDI), the Federal Ministry of Education and Research via the KB-Mining - Kompetenznetzwerk Bibliometrie project 2022-2024 (16WIK2101E).

# Data availability

All data generated or analyzed during this study, including dataset, ground truth, codebook, and all the prompts and questions for ChatGPT and BERT-based models, are openly available through enquire: tongxu@mail.cintcm.ac.cn.

# Conflicts of Interest

None declared.

# Abbreviations

BERT: Bidirectional encoder representations transformers
HB: Herb
IG: Ingredient
LLMs: Large language models
NER: Named entity recognition
NLP: Natural language processing
QA model: Question answering model
RM: Research method
SD: Study design
TCM: Traditional Chinese Medicine
TFD: The TCM formula and the Chinese patent drug
TG: Target